\documentclass[10pt,twocolumn,letterpaper]{article}

\usepackage{cvpr}
\usepackage{times}
\usepackage{epsfig}
\usepackage{graphicx}
\usepackage{amsmath}
\usepackage{amssymb}

\usepackage{subfig}
\usepackage{multirow}
\usepackage{rotating}
\usepackage{algorithm}
\usepackage{algpseudocode} 
\usepackage{amsmath}

\usepackage{tabularx, booktabs}
\usepackage{threeparttable}
\usepackage{color}
\usepackage{pifont}
\usepackage{multirow}
\usepackage{xfrac}
\usepackage[pagebackref=true,breaklinks=true,letterpaper=true,colorlinks,bookmarks=false]{hyperref}
\newcolumntype{x}[1]{>{\centering\let\newline\\\arraybackslash\hspace{0pt}}p{#1}}
\cvprfinalcopy % *** Uncomment this line for the final submission

\def\cvprPaperID{3072} % *** Enter the CVPR Paper ID here

\ifcvprfinal\fi

\begin{document}
	
	%%%%%%%%% TITLE
	%\title{Learning the Propagation and Optimization for Multi-View Stereo}
	\title{Fast-MVSNet: Sparse-to-Dense Multi-View Stereo With Learned Propagation and Gauss-Newton Refinement}
	
	\author{Zehao Yu\\
		ShanghaiTech University\\
		{\tt\small yuzh@shanghaitech.edu.cn}
		\and
		Shenghua Gao\\
		ShanghaiTech University\\
		{\tt\small gaoshh@shanghaitech.edu.cn}
	}

	\newcommand{\TODO}[1]{{\bf \textcolor{red}{[TODO: #1]}}}
	\newcommand{\REASON}[1]{{\bf \textcolor{red}{[TODO: #1]}}}
	\newcommand{\PAR}[1]{\smallskip \noindent {\bf #1~}}
	\newcommand{\PARbegin}[1]{\noindent {\bf #1~}}
	\newcommand{\red}[1]{\textcolor{red}{#1}}
	
	\maketitle
	\thispagestyle{empty}
	
	%%%%%%%%% ABSTRACT
	\begin{abstract}
Almost all previous deep learning-based multi-view stereo (MVS) approaches focus on improving reconstruction quality. Besides quality, efficiency is also a desirable feature for MVS in real scenarios. Towards this end, this paper presents a Fast-MVSNet, a novel sparse-to-dense coarse-to-fine framework, for fast and accurate depth estimation in MVS. Specifically, in our Fast-MVSNet, we first construct a sparse cost volume for learning a sparse and high-resolution depth map. Then we leverage a small-scale convolutional neural network to encode the depth dependencies for pixels within a local region to densify the sparse high-resolution depth map. At last, a simple but efficient Gauss-Newton layer is proposed to further optimize the depth map. On one hand, the high-resolution depth map, the data-adaptive propagation method and the Gauss-Newton layer jointly guarantee the effectiveness of our method. On the other hand, all modules in our Fast-MVSNet are lightweight and thus guarantee the efficiency of our approach. Besides, our approach is also memory-friendly because of the sparse depth representation. Extensive experimental results show that our method is 5$\times$ and 14$\times$ faster than Point-MVSNet and R-MVSNet, respectively, while achieving comparable or even better results on the challenging Tanks and Temples dataset as well as the DTU dataset. Code is available at \url{https://github.com/svip-lab/FastMVSNet}.
\end{abstract}

%%%%%%%%% BODY TEXT
\section{Introduction}
Multi-view stereo (MVS) aims at recovering the dense 3D structure of a scene from a set of calibrated images. It is one of the fundamental problems in computer vision and has been extensively studied for decades, because of its wide applications in 3D reconstruction, augmented reality, autonomous driving, robotics, \emph{etc}~\cite{aanaes2016large,furukawa2015multi}.

The core of MVS is the dense correspondence across images. Traditional methods usually rely on hand-crafted photo-consistency metrics (\emph{e.g.}, SSD, NCC). Designing a robust metric itself, however, is a challenging task, and thus some regularization techniques~\cite{kolmogorov2001computing,hirschmuller2007stereo} are required (\emph{e.g.}, using MRF to enforce spatial consistency~\cite{kolmogorov2001computing}). While these methods~\cite{schonberger2016pixelwise,galliani2015massively} have shown impressive results, they are still incompetent on low-textured, specular, and reflective regions where local features are not discriminative for matching. Recent work~\cite{huang2018deepmvs,yao2018mvsnet,ji2017surfacenet,kar2017learning} shows that by using Deep CNNs, the performance of MVS can be further improved. For instance, in ~\cite{yao2018mvsnet}, an MVSNet is proposed, which builds a cost volume upon CNN features and uses 3D CNNs for cost volume regularization. Such an MVSNet significantly improves the overall 3D reconstruction quality compared to traditional hand-crafted metric based methods. 

Nevertheless, all these deep learning-based methods use multi-scale 3D CNNs to predict the depth maps~\cite{yao2018mvsnet,huang2018deepmvs,im2019dpsnet} or occupancy grids~\cite{ji2017surfacenet,kar2017learning}, which are thus memory-consuming, as the memory requirement for 3D volume grows cubically. This restricts their application to high-resolution MVS. Therefore, some recent work~\cite{wang2017cnn,riegler2017octnet,yao2019recurrent,ChenPMVSNet2019ICCV} has been proposed to address this memory-intensive issue. For instance, R-MVSNet~\cite{yao2019recurrent} uses Convolutional GRU to replace 3D CNNs and thus reduces the memory requirement to quadratic, then a variational depth refinement is performed as a post-processing step to improve accuracy. Point-MVSNet~\cite{ChenPMVSNet2019ICCV} uses a coarse-to-fine strategy that first builds a relatively small 3D cost volume to predict a coarse depth map. Then a \textit{PointFlow} module is used to upsample and refine the coarse results iteratively. Although these methods avoid the memory issue in MVS and achieve state-of-the-art 3D reconstruction quality on some challenging benchmark datasets, their efficiency is still far from satisfactory. In particular, R-MVSNet~\cite{yao2019recurrent} needs 6.9 seconds to refine a depth map with size 400$\times$300 and Point-MVSNet~\cite{ChenPMVSNet2019ICCV} uses around 3 seconds to refine a depth map of the size 640$\times$480, which prohibits their application in large-scale scenarios. Besides 3D reconstruction quality, efficiency is also a desirable feature for MVS in real scenarios, which thus motivates us to work towards improving the efficiency of deep learning-based MVS methods.

Our observation is that a high-resolution depth map contains finer details, which would benefit the overall reconstruction. Directly predicting a high-resolution depth map from a 3D cost volume is, however, computationally expensive and memory-intensive. By contrast, a low-resolution depth can be predicted at much lower cost but with much fewer details. As a compromise, we propose to predict a sparse high-resolution depth map with low memory consumption first and then do the depth propagation to enrich the details with the reference image as a guidance. For depth propagation, motivated by the joint bilateral upsampling~\cite{kopf2007joint}, we propose learning a small-scale convolution neural network to encode depth dependencies for pixels within a local region to densify the sparse high-resolution depth map. We further propose using a simple and fast Gauss-Newton layer, which takes deep CNN features of multi-view images and the coarse high-resolution depth map as inputs, to refine the dense high-resolution depth map. It is worth noting that all the modules we used are lightweight and small-scale and the resulting framework can be implemented efficiently. Meanwhile, all these modules are differentiable, and thus can be trained in an end-to-end manner. We therefore term our sparse-to-dense coarse-to-fine solution as Fast-MVSNet.

In summary, our contributions are as follows: i) We propose a novel sparse-to-dense coarse-to-fine framework for MVS, where the sparse-to-dense strategy guarantees the efficiency of our approach, and the coarse-to-fine strategy guarantees the effectiveness of our approach ; ii) We propose learning the depth dependencies for pixels within a local region with a small-scale convolutional neural network from the original image and use it to densify the sparse depth map. Meanwhile, such a network is motivated by joint bilateral upsampling. Thus the depth propagation procedure is explainable; iii) A differentiable Gauss-Newton layer is proposed to optimize the depth map, which enables our Fast-MVSNet to be end-to-end learnable; iv) Extensive experiments show that our method achieves better or comparable reconstruction results compared to other state-of-the-art methods while being much more efficient and memory-friendly. In particular, our method is 14$\times$ faster than R-MVSNet~\cite{yao2019recurrent} and 5$\times$ faster than Point-MVSNet~\cite{ChenPMVSNet2019ICCV}.

\section{Related Work}
\begin{figure*}[t]
	\centering
	\includegraphics[width=1.0\textwidth]{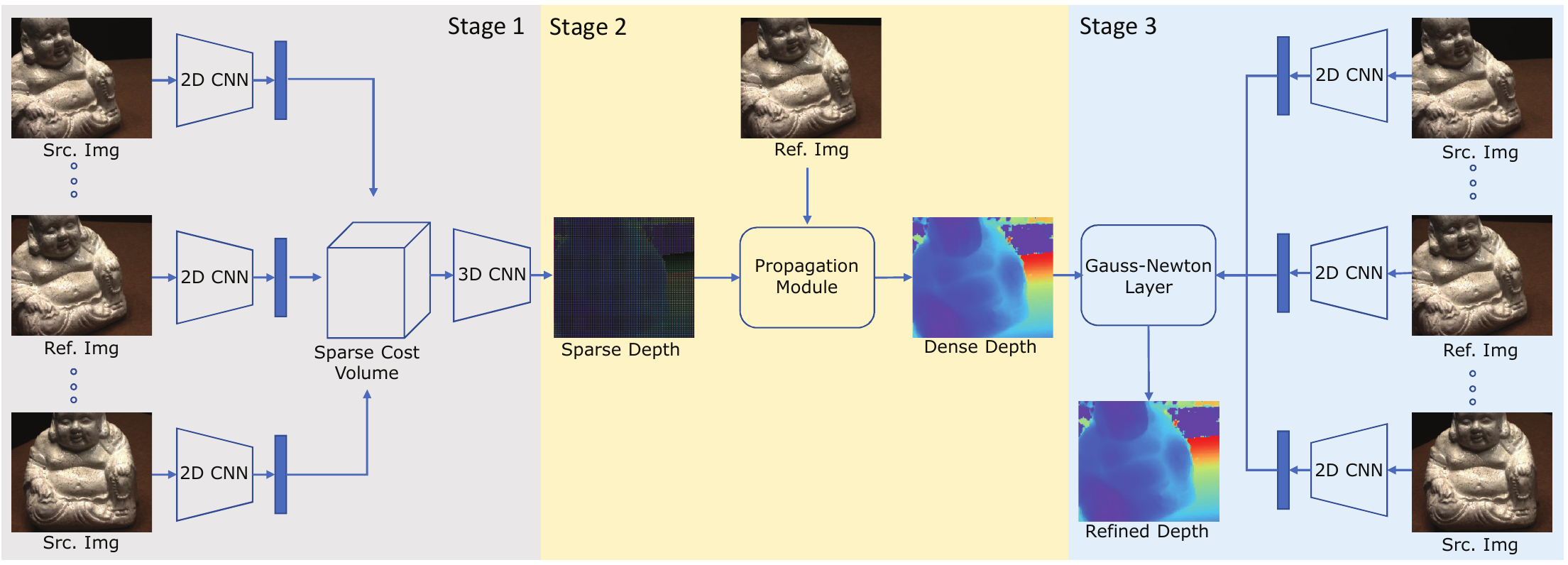}
	\caption{Network architecure of the Fast-MVSNet. In the \textbf{first stage}, we construct a sparse cost volume upon 2D CNN features and predict a sparse high-resolution depth map using a 3D CNN. In the \textbf{second stage}, we design a simple but efficient network to propagate the sparse depth map to a dense depth map. In the \textbf{third stage}, we propose a differentiable Gauss-Newton layer to further refine the depth map.\label{fig:pipeline}}
%	\vspace{-3mm}
\end{figure*}

\subsection{Multi-view stereo reconstruction}
Modern MVS algorithms usually use the following output scene representation: volume~\cite{ji2017surfacenet,kar2017learning,kutulakos2000theory,seitz1999photorealistic}, point cloud~\cite{lhuillier2005quasi,goesele2007multi,furukawa2010accurate_dense} or depth maps~\cite{tola2012efficient,schonberger2016pixelwise,galliani2015massively,yao2018mvsnet,yao2019recurrent}. In particular, volumetric based methods discretize 3D space into regular grids and decide whether a voxel is near the surface. This is of high memory consumption, however, and is not scalable to large-scale scenarios. Point cloud based methods~\cite{lhuillier2005quasi,furukawa2010accurate_dense} usually start from a sparse set of matched keypoints and use some propagation strategy to densify the point cloud. These methods, however, are difficult to be paralled as the propagation is proceeded sequentially. While depth map can be regarded as a special case of point cloud representation, (\eg, pixel-wise point cloud), it reduces the reconstruction into the problem of per-view depth map estimation. Further, one can easily fuse the depth maps to point cloud~\cite{merrell2007real} or volumetric reconstructions~\cite{newcombe2011kinectfusion}. In this work, we also use the depth map representation. It is worth noting that our method shares some similarity with point cloud based methods where we start with a sparse depth map and learn to propagate the sparse depth map to a dense one with the help of reference image.

\subsection{Learning-based MVS}
Recently, with the power of representation learning of Deep CNNs, some researchers have proposed learning better patch representations, matchings and regularizations, demonstrating great success. In ~\cite{han2015matchnet,zbontar2016stereo,hartmann2017learned}, researchers propose learning a similarity measurement between small image patches for matching cost computation. SurfaceNet~\cite{ji2017surfacenet} and DeepMVS~\cite{huang2018deepmvs} constuct a cost volume using multi-view images and use CNNs to learn the regularization of that cost volume. Yao~\etal~\cite{yao2018mvsnet} propose an end-to-end MVS architecture that builds a cost volume upon CNN features and learns the cost volume regularization also with CNNs. However, the memory consumption for 3D cost volume grows cubically. Therefore, R-MVSNet~\cite{yao2019recurrent} proposes using Convolutional GRU for cost volume regularization and thus avoid using memory-intensive 3D CNNs. By contrast, Point-MVSNet~\cite{ChenPMVSNet2019ICCV} uses a coarse-to-fine stategy that first predicts a low resolution depth map and iteratively upsamples and refines the depth map. While these method have shown impressive results, their efficiency is still far from satisfactory. Our work is mostly related to Point-MVSNet~\cite{ChenPMVSNet2019ICCV} as we also use a coarse-to-fine strategy. Instead of using a time-consuming strategy to sample depth hypotheses for refinement, however, we learn to directly optimize the depth map with a differentiable Gauss-Newton layer, which is efficient and ensures our network can be trained in an end-to-end manner. 

\subsection{Depth map upsampling and propagation}
Upsampling and propagation are ubiquitous tools in computer vision as we typically compute a low resolution result with low computational cost and interploate the result to obtain a high resolution result. Simple upsampling methods such as nearest neighbour and bilinear interpolation, however, subsequently suffer from over smoothing around image edges. Instead, by using high resolution image as a guidance, joint bilateral upsampling~\cite{kopf2007joint,barron2016fast} can preserve edge characteristics. 
Xu~\etal~\cite{xu2019multi} further propose using multi-view geometric consistency as a guidance for depth map upsampling, while Wei~\etal~\cite{wei2019joint} extend joint bilateral upsamling to incorporate surface normal information. These methods, however, rely on hand-crafted strategy and their kernel parameters need to be manually tuned. Unlike these methods, we propose learning the propagation of our sparse depth map with image guidance and show that by incorporating a learnable propagation module, reconstruction results can be further improved. 

\subsection{Learning-based optimization}
Some recent work has been proposed to learn the optimization of nonlinear least square objective functions by utilizing the differentiable nature of iterative optimization algorithms. 
These optimization algorithms are unrolled for a fixed number of iterations, and each iteration is implemented as a layer in a neural network. In ~\cite{Clark_2018_ECCV}, an LSTM~\cite{hochreiter1997long} is used to model the Levenberg-Marquardt (LM) algorithm and predicts the update at each step directly. In ~\cite{tang2018ba}, Tang~\etal propose a differentiable LM algorithm by learning to predict the damping factor of standard LM algorithm, while Lv~\etal~\cite{lv2019taking} use learnable modules to replace multiple components of the inverse compositional algorithm~\cite{baker2004lucas}. Unlike these methods, CodeSLAM~\cite{bloesch2018codeslam} and SceneCode~\cite{Zhi_2019_CVPR} learn a compact representation (\ie, \textit{code}) of scene for later optimization, while Stumberg~\etal~\cite{von2019gn} propose a Gauss-Newton loss to learn robust representation of images under different weather conditions. Our method is particularly inspired by this line of work where we propose a differentiable Gauss-Newton layer for efficient depth map refinement but our method is not restricted to scene dependent depth basis~\cite{tang2018ba} or learned code~\cite{bloesch2018codeslam,Zhi_2019_CVPR}.
\section{Method}

%Overview
Our goal is to design an effective and efficient framework for MVS. Following recent successes~\cite{huang2018deepmvs,yao2018mvsnet,yao2019recurrent,ChenPMVSNet2019ICCV,im2019dpsnet}, we use per-view depth map as scene representation for its flexibility and scalability. That is to estimate a depth map for a reference image $I_0$ given a set of neighboring source images $\{I_i\}_{i=1}^N$.

To this end, we propose a Fast-MVSNet, an efficient MVS framework that ultizes a sparse-to-dense coarse-to-fine strategy for depth map estimation. Specifically, we \textbf{first} estimate a sparse high-resolution depth map such that existing MVS methods can be applied at a lower cost (\ie, less computational cost and less memory consumption). \textbf{Then} we design a simple but efficient propagation module to propagate the sparse depth map to a dense depth map. \textbf{Finally}, a differentiable Gauss-Newton layer is proposed to further optimize the depth map for sub-pixel accuracy. The overall pipeline of our method is shown in Figure~\ref{fig:pipeline}. Next, we will introduce each component of our method in details.

\subsection{Sparse high-resolution depth map prediction}

\begin{figure}[t]
	\centering
	\footnotesize
	\setlength{\abovecaptionskip}{0.1cm}
	\setlength{\belowcaptionskip}{-0.3cm}
	\includegraphics[width=1.0\columnwidth]{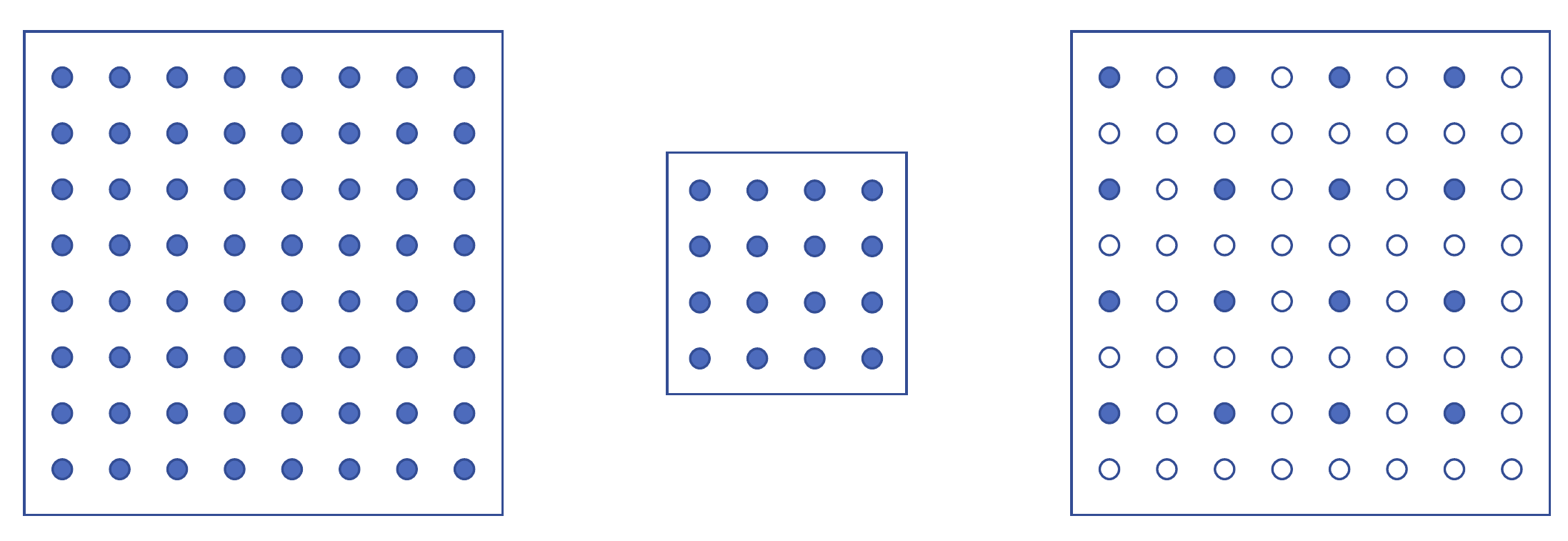}
	\begin{tabular}{x{0.28\columnwidth}x{0.3\columnwidth}x{0.28\columnwidth}}
		(a) & (b) & (c)\\
	\end{tabular}
	
	\caption{Initialization of depth map. (a) MVSNet~\cite{yao2018mvsnet} and R-MVSNet~\cite{yao2019recurrent}. (b) PointMVSNet~\cite{ChenPMVSNet2019ICCV}. (c) Ours. Unlike other methods, we estimate a sparse high-resolution depth map considering efficiency and quality. \label{fig:init}}	
	\vspace{-1mm}
	
\end{figure}

Our first step is to estimate a sparse high-resolution depth map for the reference image $I_0$. Figure~\ref{fig:init} shows the key differences between our sparse depth map representation and depth maps in other methods. We estimate a sparse high-resolution depth map with low memory and computation costs while other methods either estimate a high-resolution depth  map~\cite{huang2018deepmvs,yao2018mvsnet} with high memory costs or a low-resolution depth map~\cite{ChenPMVSNet2019ICCV} without fine details. We argue that our sparse high-resolution representation is more adequate than a low-resolution representation because: i) training with a low-resolution depth map requires downsampling the ground-truth depth map accordingly. If we downsample the ground-truth depth map with the nearest neighbour method, then the low-resolution representation is the same as our sparse high-resolution representation. In this case, however, the resulting depth map is not well aligned with the extracted low-resolution feature map. If we use bilinear interpolation for downsampling, it will cause artifacts around regions with depth discontinuities; ii) fine details are lost in the low-resolution depth map. Recovering a high-resolution depth map with fine details from a low-resolution one requires non-trival and complicated upscaling methods~\cite{dosovitskiy2015flownet}.%\TODO{figure}

To predict our sparse high-resolution depth map, we adapt the MVSNet~\cite{yao2018mvsnet} for our task. Specifically, we first use the same 8-layer 2D CNN as MVSNet to extract image features, then we build a sparse cost volume in the frustum of the reference image. Finally, we use 3D CNN to regularize the cost volume and predict a sparse depth map via differentiable argmax~\cite{yao2018mvsnet}. Our method is a general framework. While we use 3D CNN for cost volume regularization, other regularization methods such as Convolutional GRU~\cite{yao2019recurrent} are also applicable.

We highlight the differences of our cost volume with previous methods~\cite{yao2018mvsnet,ChenPMVSNet2019ICCV} as follows: i) our cost volume is of size $\frac{1}{8}H \times \frac{1}{8}W \times N \times F$, while MVSNet use a cost volume of size $\frac{1}{4}H \times \frac{1}{4}W \times N \times F$, where $N$ is the number of depth planes and $F$ is the number of feature channels; ii) MVSNet uses 256 virtual depth planes, while we use the same number of depth planes as that in  Point-MVSNet~\cite{ChenPMVSNet2019ICCV}. In particular, we use 48 and 96 virtual depth planes for training and evaluation respectively; iii) We use an 8-layer 2D CNN to extract image features with $F=32$ channels while Point-MVSNet~\cite{ChenPMVSNet2019ICCV} uses an 11-layer 2D CNN to extract image features with $F=64$ channels. As a result, the memory usage of our cost volume is $\frac{1}{2}$ of that in Point-MVSNet~\cite{ChenPMVSNet2019ICCV}.

Interestingly, due to our sparse representation, the 3D CNN acts like dilated convolutions~\cite{chen2017deeplab} with dilation 2 in the spatial domain. Thus it has the potential to incorporate larger spatial contextual information for regularization. 

\subsection{Depth map propagation}
\begin{figure}[t]
	\centering
	\includegraphics[width=1.0\columnwidth]{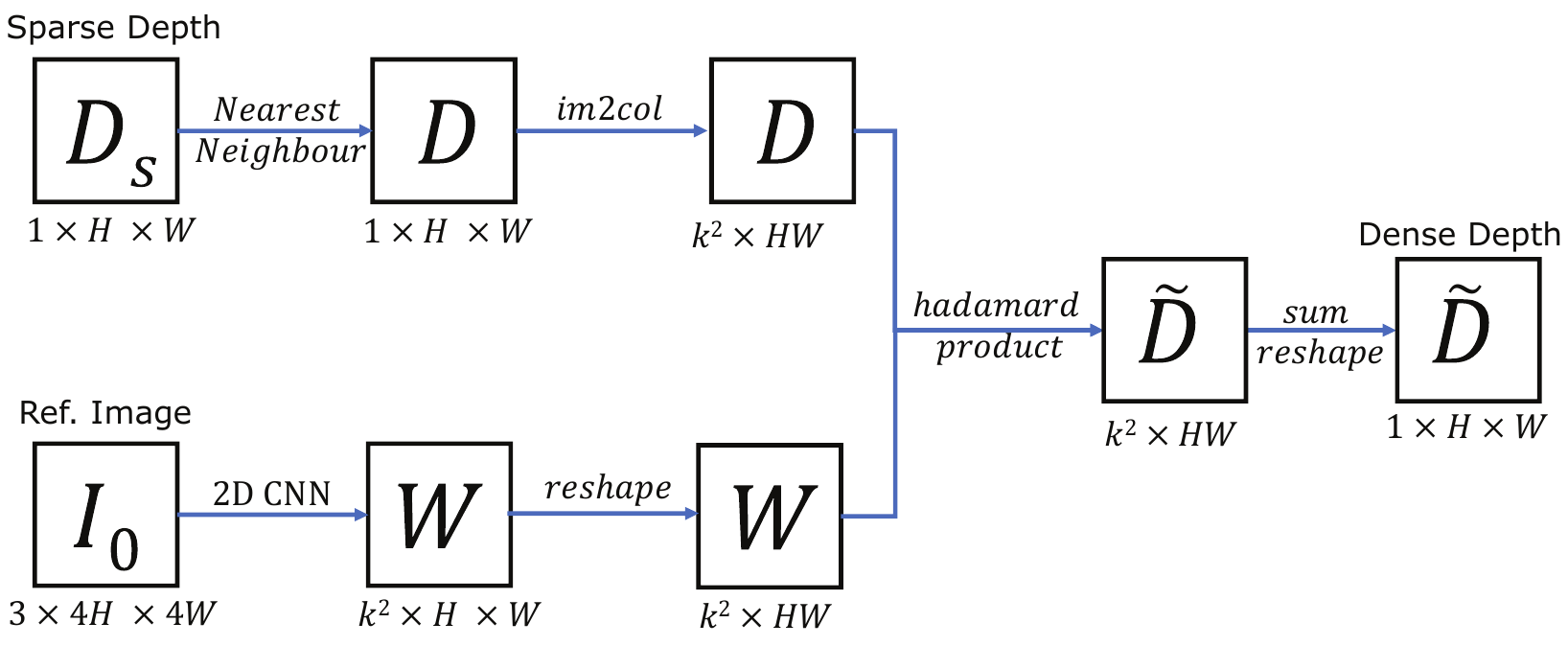}
	\caption{A diagram of the propagation module. \label{fig:propagation}}	
	\vspace{-2mm}
\end{figure}

The former step provides us a high-resolution but sparse depth map $D$. We now need to propagate the sparse depth map to obtain a dense depth map $\tilde{D}$. A simple strategy is to use nearest neighbour for this purpose. This nearest neighbor method, however, does not consider the original image information and thus may not work well around depth boundaries. Another natural choice is the joint bilateral upsampler~\cite{kopf2007joint,barron2016fast,wei2019joint} which uses the information of original high-resolution image as a guidance. Formally, it takes the following form:
\begin{equation}
\tilde{D}(p) = \frac{1}{z_p} \sum_{q\in N(p)} D(q) f(\| p - q \|) g(\| I_p - I_q \|)
\label{eq:bilateral}
\end{equation}
where $f$ is the spatial filter kernel, $g$ is the range filter kernel, $N(p)$ is the local $k\times k$ neighbour around position $p$ and $z_p$ is a normalization term. These two kernel parameters, however, may be different for diverse scenes and need to be manually tuned. 

We therefore propose replacing $f(\| p - q \|) g(\| I_p - I_q \|)$ with a weight $w_{p,q}$ and learning the weights with a simple network. Mathematically, we use the following form:
\begin{equation}
\tilde{D}(p) = \frac{1}{z_p} \sum_{q\in N(p)} D(q) \cdot w_{p,q}
\label{eq:learned_propagate}
\end{equation}
where $w_{p,q}$ is the output of a CNN and is learned in a data-driven manner. We note that while we do not explicitly account for spatial information, it is indeed implicitly encoded by the network. Further, as we predict different weights for different position $p$, our method can be viewed as a generalization of the standard bilateral upsampler that applies a fixed kernel for every position $p$.

\PARbegin{Implementation.} 
The sparse depth map $D_s$ is first propagate to a dense depth map $D$ using nearest neighbour. In parallel, a CNN takes the reference image $I_0$ as input and outputs $k\times k$ weights $W$ for each position. Finally, the propagated depth map $\tilde{D}$ is computed using Equation~\ref{eq:learned_propagate}. Note that the computation of Equation~\ref{eq:learned_propagate} can be efficiently implemented using vectorization (\ie \textit{im2col}). The details of the proposed propagation module are shown in Figure~\ref{fig:propagation}. To predict the weight $W$, we simply use the same network architecture as that in MVSNet to extract image features and append a two-layer 3$\times$3 convolutional network to predict a feature map with $k \times k$ channels. The softmax function is applied in the channel dimension for normalization. 

\subsection{Gauss-Newton refinement}
\begin{figure}[t]
	\centering
	\includegraphics[width=0.95\columnwidth]{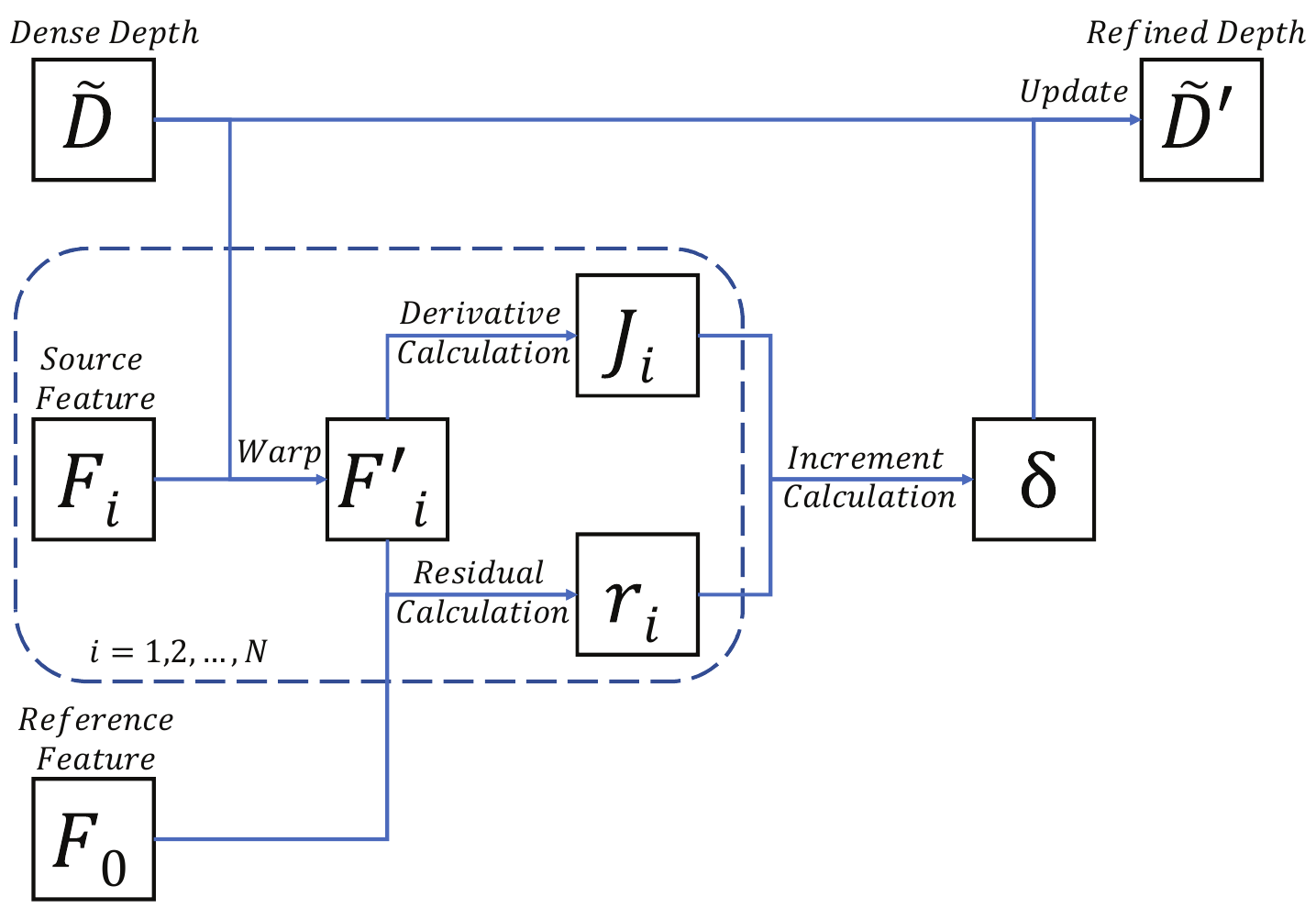}
	\caption{A diagram of the differentiable Gauss-Newton Layer. We ignore camera parameters here for simplicity. \label{fig:gn}}	
	\vspace{-2mm}
\end{figure}

As we concentrate on the efficient inference of a dense high-resolution depth map in the previous step, the accuracy of the resulting depth map is insufficient. Therefore, we propose using the Gauss-Newton algorithm to refine the depth map. While there are various methods could be used for depth map refinement~\cite{yao2019recurrent,yao2018mvsnet,ChenPMVSNet2019ICCV}, we select the Gauss-Newton algorithm for its efficiency. 

Mathematically, given a point $p$ with depth $\tilde{D}_p$ in the reference image, we aim to minimize the following error function:
\begin{equation}
E(p) = \sum_{i = 1}^N \|F_i(p'_i) - F_0(p)\|_2
\label{eq:error}
\end{equation}
where $F_i$ and $F_0$ are the deep representation of the source image $I_i$ and the reference image $I_0$, respectively, $p'_i$ is the reprojected point of $p$ in image $I_i$ and  $F_i(p)$ corresponds to the features at $p$ in $F_i$. $p'_i$ can be computed as 
\begin{equation}
p'_i = K_i (R_i R_0^{-1} ( \tilde{D}(p) K_0^{-1} p - t_0 ) + t_i )
\label{eq:reproject}
\end{equation}
where $\{K_i, R_i, t_i\}_{i=0}^N$ denote the camera intrinsics, rotations and translations of corresponding images.

We apply the Gauss-Newton algorithm to minimize $E(p)$. Specifically, starting with an initial depth $\tilde{D}_p$, we compute the residual $r_i(p)$ of $p$ for each source image $I_i$:
\begin{equation}
r_i(p) = F_i(p'_i) - F_0(p) 
\label{eq:residual}
\end{equation}
Then for each residual $r_i(p)$, we compute their first order derivative with respect to $\tilde{D}(p)$ as:
\begin{equation}
%J_i(p) = \frac{\partial r_i(p)}{\partial \tilde{D}_p} = \frac{\partial F_i(p_i')}{\partial p_i'} \cdot \frac{\partial p_i'}{\partial \tilde{D}_p} = \frac{\partial F_i(p_i')}{\partial p_i'} \cdot K_i R_i R_0^{-1} K_0^{-1} p
%J_i(p) = \frac{\partial r_i(p)}{\partial \tilde{D}_p} = \frac{\partial F_i(p_i')}{\partial p_i'} \cdot \frac{\partial p_i'}{\partial \tilde{D}_p} 
J_i(p) = \frac{\partial F_i(p_i')}{\partial p_i'} \cdot \frac{\partial p_i'}{\partial \tilde{D}(p)} 
\label{eq:jacobian}
\end{equation}
Finally, we can obtain the increment $\delta$ to the current depth as:
\begin{equation}
\delta = -(J^T J)^{-1} J^T r
\label{eq:soft_assignment}
\end{equation}
where $J$ is the stack of jacobians $\{J_i(p)\}_{i=1}^N$, and $r$ is the stack of residual vectors $\{r_i(p)\}_{i=1}^N$. Therefore, the refined depth is:
\begin{equation}
\tilde{D}'(p) = \tilde{D}(p) + \delta .
\label{eq:soft_assignment}
\end{equation}

Further, the Gauss-Newton algorithm is naturally differentiable and can be implemented as a layer in a neural network without additional learnable parameters. As shown in Figure~\ref{fig:gn}, the Gauss-Newton layer takes multi-view image features, camera parameters and an initial depth map as input, then outputs a refined depth map. The overall network can be trained in an end-to-end manner. Therefore, it can learn suitable features for efficient optimization~\cite{tang2018ba}. We find that it converges quickly with only a single step of update.
%While the Gauss-Newton algorithm is an iterative method, we found that it converges quickly with a single step of update.
Note that our Gauss-Newton layer is different with the refinement of R-MVSNet which use gradient decent to optimize hard-crafted photo-consistency metric, whereas we intergrate the optimization in training. Furthermore, as we do not need to sample depth hypotheses, our method is more efficient and memory-friendly compared to Point-MVSNet~\cite{ChenPMVSNet2019ICCV}.

\subsection{Training Loss}
Following previous methods~\cite{yao2018mvsnet,ChenPMVSNet2019ICCV}, we use the mean absolute difference between the estimated depth map and ground truth depth map as our training loss. Both the initial depth map $\tilde{D}$ and the refined depth map $\tilde{D}'$ are included in our trianing loss:
\begin{equation}
Loss = \sum_{p \in \textbf{p}_{valid}} \| \tilde{D}(p) - \hat{D}(p)\| + \lambda \cdot \| \tilde{D}'(p) - \hat{D}(p)\|
\label{eq:loss}
\end{equation}
where $\hat{D}$ is the ground truth depth map, $\textbf{p}_{valid}$ denotes the set of valid ground truth depths and $\lambda$ is the weight that balances the two losses. We set $\lambda$ to be $1.0$ in all the experiments.

\section{Experiments}
\begin{figure*}[t!]
	\setlength{\abovecaptionskip}{0.05cm}
	\setlength{\belowcaptionskip}{-0.4cm}
	\centering
	\footnotesize
	\includegraphics[width=0.9\textwidth]{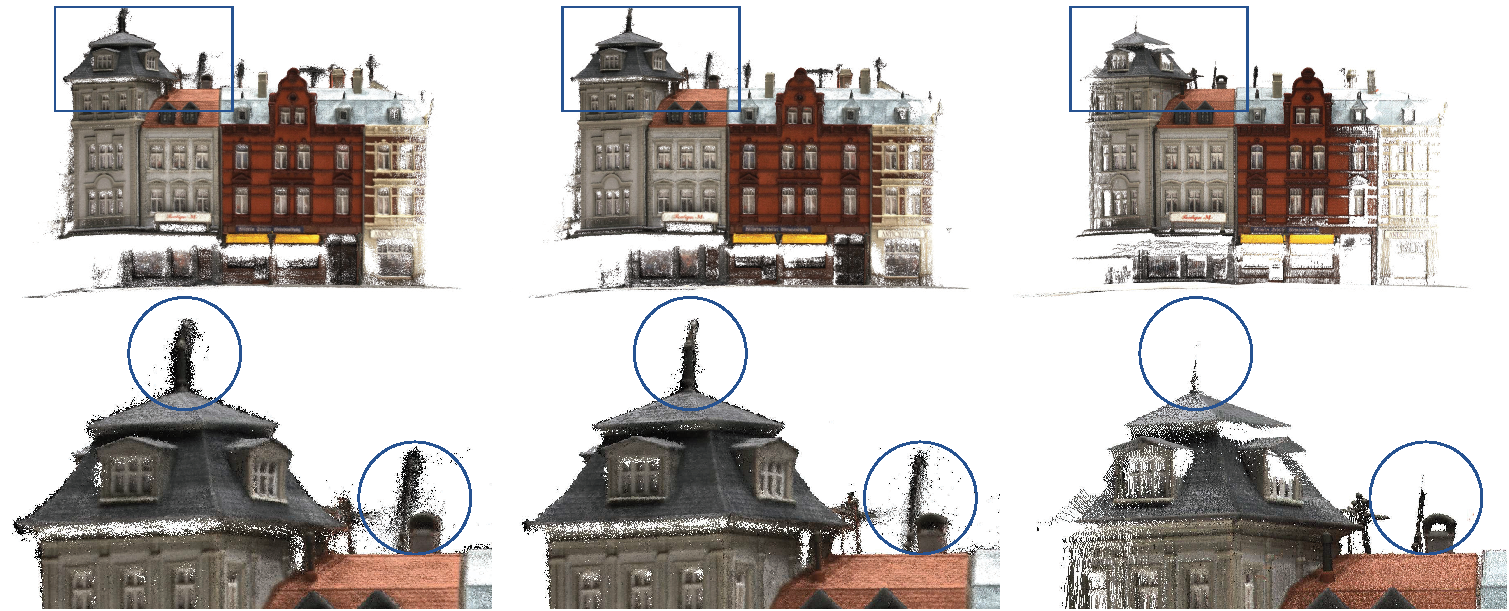}
	\begin{tabular}{x{0.3\linewidth}x{0.3\linewidth}x{0.3\linewidth}}
		Point-MVSNet~\cite{ChenPMVSNet2019ICCV} & Ours & Ground Truth\\
	\end{tabular}
	\caption{Qualitative results of \textit{scan9} of the DTU dataset. Top: Whole point cloud. Bottom: Zoomed local region. We use the same point cloud fusion parameters with Point-MVSNet. As shown in blue circle region, our reconstruction contains less noise around fine detailed structures, which demonstrates the effectiveness of our method.\label{fig:dtu:visual}}
\end{figure*}

\begin{table}[t]
	\centering
	\setlength{\abovecaptionskip}{0.05cm}
	\setlength{\belowcaptionskip}{-0.3cm}
	\footnotesize
	\begin{tabular}{r|ccc}
		\toprule
		& Acc. (mm)    & Comp. (mm) & Overall (mm) \\
		\midrule
		Camp~\cite{campbell2008using_multiple}   &    0.835&    0.554&    0.695 \\
		Furu~\cite{furukawa2010accurate_dense}   &0.613    &    0.941&0.777    \\
		Tola~\cite{tola2012EfficientLargescaleMultiview}  &0.342    & 1.190    &0.766     \\
		Gipuma~\cite{galliani2015massively}  &\textbf{0.283}    &0.873    & 0.578 \\				PU-Net~\cite{yu2018punet}  &1.220    &0.667    & 0.943 \\
		SurfaceNet~\cite{ji2017surfacenet} &0.450    &1.040     & 0.745\\
		MVSNet~\cite{yao2018mvsnet}  & 0.396 & 0.527    & 0.462 \\
		R-MVSNet~\cite{yao2019recurrent}    & 0.385 &  0.452 & 0.417    \\
		PointMVSNet~\cite{ChenPMVSNet2019ICCV}    & 0.361 &  0.421 & 0.391    \\
		\midrule
		Ours    & 0.336 &  \textbf{0.403} & \textbf{0.370}    \\
		\bottomrule      
	\end{tabular}
	\caption{Quantitative results of reconstruction quality on the DTU evaluation data~\cite{aanaes2016large}. Our method outperforms all methods in terms of reconstruction completeness and overall quality. }
	\label{tab:res-dtu}
	\vspace{-2mm}
\end{table}

\subsection{The DTU dataset}
The DTU dataset~\cite{aanaes2016large} is a large scale MVS dataset, which contains 80 scenes with large diversity. Each scene is captured at 49 or 64 precise camera positions with 7 different lighting conditions. The dataset provides reference models which are acquired by an accurate structured light scanner along with high-resolution RGB images. We use the same training, validation and evaluation sets as that in other learning based methods~\cite{ji2017surfacenet,yao2018mvsnet,yao2019recurrent,ChenPMVSNet2019ICCV}.
 
\subsection{Implementation Details}

\PARbegin{Training.} 
We use the training data generated by MVSNet~\cite{yao2018mvsnet}. The point cloud provided by the DTU dataset are used to reconstruct mesh surfaces which are then used to render depth maps for training. We implement our model with PyTorch~\cite{Paszke2017PyTorch}. We set the resolution of input image to 640$\times$512 and the number of views $N$ to 3. To choose source images for training, the same view selection strategy as MVSNet~\cite{yao2018mvsnet} is used. We set the number of depth planes $D=48$ in sparse depth map prediction where depth hypotheses are uniformly sampled from 425$mm$ to 921$mm$. Following PointMVSNet~\cite{ChenPMVSNet2019ICCV}, we use the RMSProp optimizer with initial learning rate 0.0005 and decrease the learning rate by 0.9 every 2 epochs. The batch size is set to 16 on 4 NVIDIA GTX 2080Ti GPU devices. We first pretrain the sparse depth map prediction module and propagation module for 4 epochs. Then the overall model is trained end-to-end for another 12 epochs. Details of network architecture are described in supplementary material.

\PARbegin{Testing.}
After the propagation of sparse depth map, we get a dense depth map of size $1 \times H \times W$. For a fair comparison with Point-MVSNet~\cite{ChenPMVSNet2019ICCV},  we upsample the depth map to $2 \times H \times W$ with nearest neighbor before Gauss-Newton refinement. We use $N=5$ images with resolution 1280$\times$960 as input and we set the number of depth planes $D=96$. We first predict a depth map for each reference image and then use the post processing provided by ~\cite{yao2018mvsnet} to fuse the predicted depth maps into point cloud. The same parameters for depth map fusion as Point-MVSNet~\cite{ChenPMVSNet2019ICCV} are used unless otherwise specified.

%\PARbegin{Metrics.}
%For quantitative evaluation, we use the officially prescribed metrics in the DTU dataset~\cite{aanaes2016large}: \textit{Accuracy}, \textit{Completeness} and \textit{Overall}. \textit{Accuracy} measures how close the reconstructed points to the ground truth are. It is defined as the mean distance from the reconstructed points to the ground truth. \textit{Completeness} measures to what extent the ground truth points are recovered and is defined as the mean distance from the ground truth points to the reconstructed points. \textit{Overall} is the mean of \textit{Accuracy} and \textit{Completeness}. It measures the overall reconstruction quality. We use the MATLAB code provided by the DTU dataset to compute these evaluation metrics.

\subsection{Results on the DTU dataset}

\begin{figure*}[t!]
	\setlength{\abovecaptionskip}{0.1cm}
	\setlength{\belowcaptionskip}{-0.3cm}
	\centering
	\footnotesize
	\includegraphics[width=0.9\textwidth]{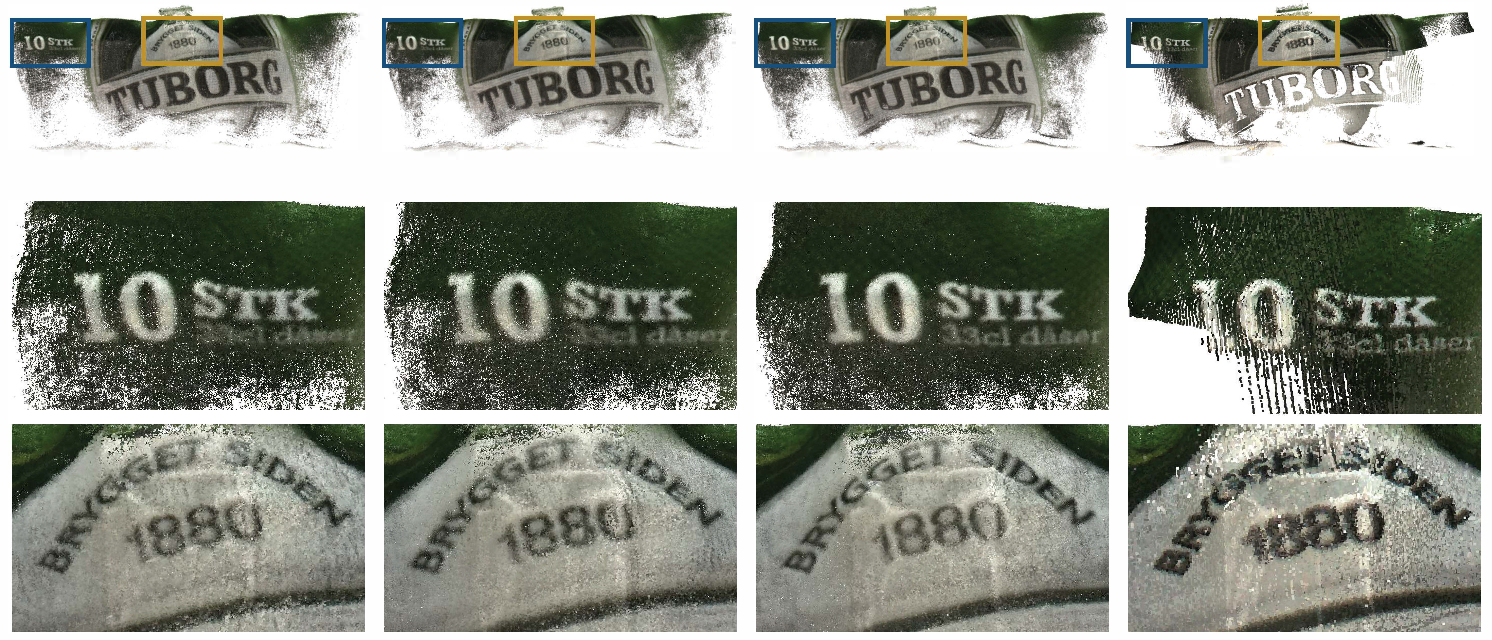}
	\begin{tabular}{x{0.205\linewidth}x{0.20\linewidth}x{0.205\linewidth}x{0.20\linewidth}}
		(a): Sparse High-Resolution & (b): (a) $+$ Propagation Module & (c): (b) $+$ Gauss-Newton Layer & Ground Truth\\
	\end{tabular}
	\caption{Qualitative results of \textit{scan12} of the DTU dataset. Top: whole point cloud. Middle and bottom: zoomed local region of rectangle. Reconstruction results become denser and more detailed when gradually adding propagation module and Gauss-Newton layer (see text region). \label{fig:dtu:ab}}
\end{figure*}

\begin{table*}[!t]
	\setlength{\abovecaptionskip}{0.00cm}
	\setlength{\belowcaptionskip}{-0.4cm}
	\centering
	\footnotesize
	\begin{tabular}{@{}c|ccc|c|cc@{}}
		\toprule
		& Acc. (mm) & Comp. (mm) & Overall (mm) & Depth Map Res. & GPU Mem. (GB) & Runtime (s)  \\ \midrule
		MVSNet\cite{yao2018mvsnet} & 0.456 & 0.646 & 0.551 & 288$\times$216 & 10.8 & 1.05\\
		R-MVSNet\cite{yao2018mvsnet} & 0.385 &  0.452 & 0.417 & 400$\times$300 & 6.7 & 9.1\\
		Point-MVSNet~\cite{ChenPMVSNet2019ICCV} & 0.361 & 0.421 & 0.391 & \textbf{640$\times$480} & 8.7 & 3.35 \\ \midrule
		Ours & \textbf{0.336} &  \textbf{0.403} & \textbf{0.370} & \textbf{640$\times$480} & \textbf{5.3} & \textbf{0.6} \\
		\bottomrule
	\end{tabular}
	\caption{Comparison results measured by reconstruction quality, depth map resolution, GPU memory requirements and runtime on the DTU evaluation set. The result of MVSNet~\cite{yao2018mvsnet} is quoted from Point-MVSNet~\cite{ChenPMVSNet2019ICCV}. Due to the GPU memory limitation, the resolution of MVSNet~\cite{yao2018mvsnet} is decreased to 1152$\times$864$\times$192. Our method outperforms all methods in terms of all evaluation metrics while being more efficient and more memory-friendly. }
	\label{tab:runtime}
\end{table*}

We compare our method with both traditional methods and recent learning based methods. The quantitative results are shown in Table~\ref{tab:res-dtu}. While Gipuma~\cite{galliani2015massively} achieves the best performance in terms of \textit{Accuracy}, our method outperforms all competing methods in both \textit{Completeness} and \textit{Overall} quality. Figure~\ref{fig:dtu:visual} shows the qualitative comparison with the results of Point-MVSNet. Our reconstruction is cleaner around fine detailed structures, which validates the effectiveness of our method.

We further demonstrate the efficiency and the effectiveness of the proposed method by comparing the reconstruction quality, depth map resolution, GPU memory requirements and runtime with state-of-the-art methods in Table~\ref{tab:runtime}. For a fair comparison with Point-MVSNet~\cite{ChenPMVSNet2019ICCV}, the runtime is measured on an NVIDIA GTX 1080Ti GPU. As shown in Table~\ref{tab:runtime}, our method outperforms all methods in terms of all evaluation metrics while being more efficient and more memory-friendly. In particular, our method is about 2$\times$ faster than MVSNet~\cite{yao2018mvsnet}, 14$\times$ faster than R-MVSNet~\cite{yao2019recurrent} and 5$\times$ faster than Point-MVSNet~\cite{ChenPMVSNet2019ICCV}. 

\begin{figure*}[t]
	\setlength{\abovecaptionskip}{0.05cm}
	\setlength{\belowcaptionskip}{-0.2cm}
	\setlength{\tabcolsep}{0.1em}
	\renewcommand{\arraystretch}{0.1}
	\footnotesize
	\centering
	\includegraphics[width=0.9\textwidth]{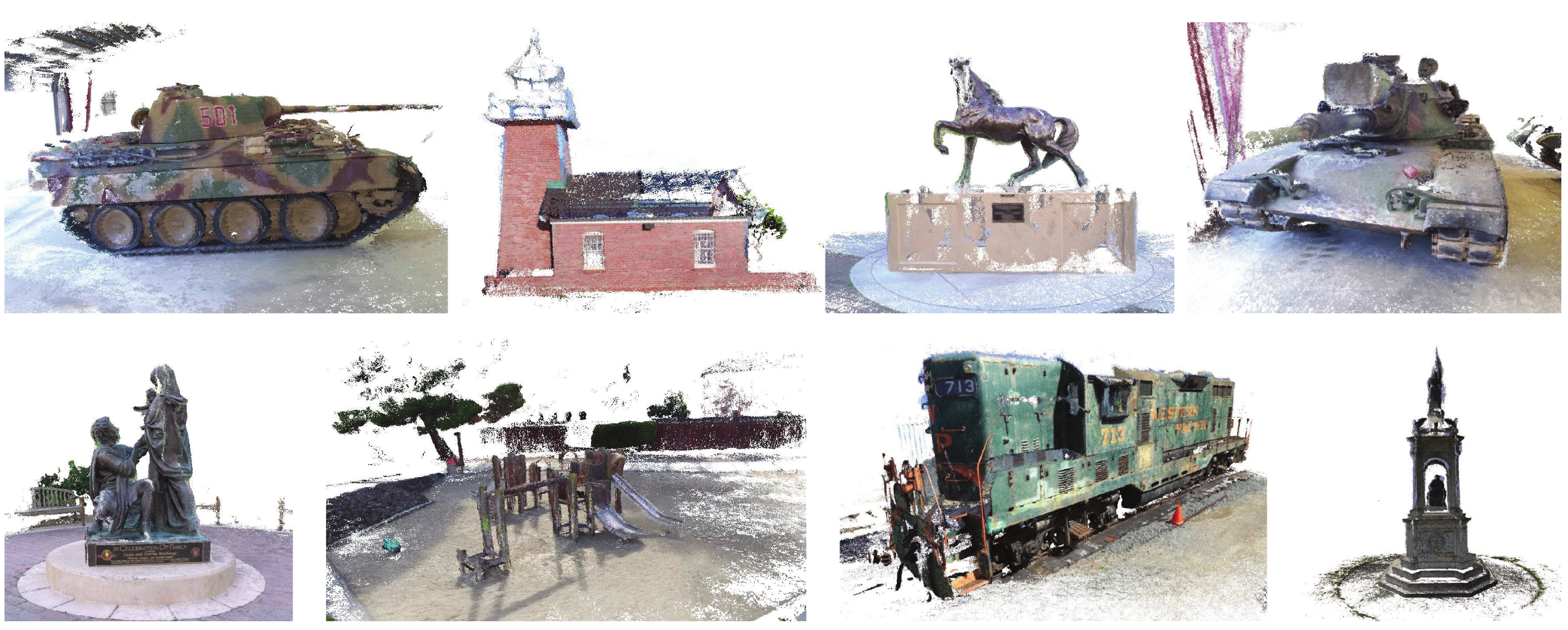}
	\caption{Reconstruction results of \textit{intermediate set} in the \textit{Tank and Temples} dataset~\cite{Knapitsch2017}. Our method can reconstruct dense and visually appealing complex scenes. \label{fig:tt:visual}}
	%	\vspace{-3mm}
\end{figure*}

\begin{table*}[!t]
	\setlength{\abovecaptionskip}{0.05cm}
	\setlength{\belowcaptionskip}{-0.5cm}
	\centering
	\footnotesize
	\begin{tabular}{c|ccccccccc}
		\toprule
		& Mean    		& Family		& Francis & Horse 		& Lighthouse 	& M60   	 & Panther 		& Playground 	& Train \\ \hline
		MVSNet~\cite{yao2018mvsnet}	 & 43.48   & 55.99	& 28.55	 & 25.07 & \textbf{50.79} & \textbf{53.96} & \textbf{50.86} & 47.90 & 34.69 \\
		R-MVSNet~\cite{yao2019recurrent} & \textbf{48.40} & \textbf{69.96} & \textbf{46.65} & 32.59 & 42.95 & 51.88 & 48.80 & 52.00 & 42.38 \\
		Point-MVSNet~\cite{ChenPMVSNet2019ICCV} & 48.27	& 61.79 & 41.15 & 34.20 & \textbf{50.79} & 51.97 & 50.85 & 52.38 & \textbf{43.06} \\
		\midrule
		Ours     & 47.39   & 65.18	& 39.59	 & \textbf{34.98} & 47.81 & 49.16 & 46.20 & \textbf{53.27} & 42.91 \\
		\bottomrule
	\end{tabular}
	\caption{Evaluation results on \textit{Tanks and Temples} benchmark~\cite{Knapitsch2017}. We achieve comparable results with state-of-the-art methods. }
	\label{tab:tt}
\end{table*}

\subsection{Ablation study}
Due to space limitations, we refer readers to supplementary material for additional ablation studies, including Gauss-Newton refinements with more iterations and depth map fusion with different parameters.

\begin{table}[!t]
	\setlength{\abovecaptionskip}{0.05cm}
	\setlength{\belowcaptionskip}{-0.5cm}
	\centering
	\footnotesize
	\begin{tabular}{c|cccc}
		\toprule
		method & Acc. & Comp. & Overall \\
		\midrule
		low res. &   0.517 &  0.557 & 0.537 \\
		sparse high res. &  0.394 & 0.478 & 0.436 \\
		sparse high res. + prop. & 0.370 & 0.448 & 0.409 \\
		sparse high res. + prop. + GN & \textbf{0.336} &  \textbf{0.403} & \textbf{0.370} \\
		\midrule
		low res. + \textit{PointFlow}~\cite{ChenPMVSNet2019ICCV} & 0.361 &  0.421 & 0.391 \\
		low res. + GN. & 0.376 & 0.417 & 0.396 \\
		\bottomrule
	\end{tabular}
	\caption{Ablation study on the DTU evaluation dataset, which demonstrates the effectiveness of different components of our method. Low res. denotes low-resolution depth map, sparse high res. denotes our sparse high-resolution depth map, prop. denotes propagation module and GN denotes Gauss-Newton refinement.}
	\label{tab:ablation}
\end{table}

%In this section, we conduct abalation study to verify the design choice of our method.
\PARbegin{Effectiveness of the sparse high-resolution depth map.} To evaluate the effectiveness of our sparse high-resolution depth map representation, we compare the reconstruction results with the low-resolution depth map representation in Table~\ref{tab:ablation}. For a fair comparison, both the low-resolution depth map and our sparse high-resolution depth map are upsampled to the 640$\times$ 480 with nearest neighbor. As shown in the first two rows in Table~\ref{tab:ablation}, our sparse high-resolution depth map achieve better results. 

\PARbegin{Effectiveness of propagation module.} To evaluate the effectiveness of our learned propagation module, we show the results with or without the propagation module in the second and third row in Table~\ref{tab:ablation}, showing the propagation module can further improved the reconstruction results.

\PARbegin{Effectiveness of the Gauss-Newton Refinement.} We compare the results with or without the Gauss-Newton refinement in the third and fouth row in Table~\ref{tab:ablation}. With the Gauss-Newton refinement, the relative improvement of \textit{Overall} reconstruction quality is $9.5\%$ (from 0.409 to 0.370), showing the effectiveness of our Gauss-Newton Refinement. 

\PARbegin{Efficiency of the Gauss-Newton Refinement.}
Both our Gauss-Newton layer and the \textit{PointFlow} module proposed in Point-MVSNet~\cite{ChenPMVSNet2019ICCV} aim to refine a coarse depth map. \textit{PointFlow} uses a hypothesis testing strategy that first samples a set of hypotheses (around the current depth prediction) and uses a network to select a better hypothesis via weighted average among all hypotheses. In the contrary, we formulate the depth map refinement as an optimization problem and intergate the optimization into an end-to-end framework. Compared with hypotheses sampling solution in Point-MVSNet, our formulation is simple and more efficient. 

In order to evaluate the efficiency of the proposed Gauss-Newton layer, we replace the \textit{PointFlow} Module of Point-MVSNet with our differentiable Gauss-Newton layer and train the network from scratch. The comparison of reconstruction results are shown in the last two rows in Table~\ref{tab:ablation}. We achieve comparable results with Point-MVSNet while our method is 5$\times$ faster. Further, as we directly optimize the depth instead of sampling possible depth hypotheses, our method is more memory-friendly and does not need to adopt a divide and conquer strategy to refine a high-resolution depth map (\eg 640$\times$480).

We show the comparison of recontruction results when adding different components of our method in Figure~\ref{fig:dtu:ab}. The results become denser and contains much finer details especilly in the text region.

\subsection{Generalization}
To evaluate the generalizability of our proposed method, we test it on the large scale \textit{Tanks and Temples} dataset~\cite{Knapitsch2017}. We use the model trained on the DTU dataset without fine-tuning for testing. We use $N=5$ images with resolution 1920$\times$1056 as input. We set the number of depth planes $D=96$. We use the camera parameters provided by MVSNet~\cite{yao2018mvsnet} for a fair comparison. The evaluation results are shown in Table~\ref{tab:runtime}. We achieve comparable results with state-of-the-art methods, which demonstrates the generalizability of the proposed method. Qualitative results are shown in Figure~\ref{fig:tt:visual}. Our reconstruction is dense and visually apearling. 

\section{Conclusion}
We propose Fast-MVSNet as an efficient MVS solution, which leverages a sparse-to-dense coarse-to-fine strategy. We first estimate a sparse high-resolution depth map at a lower costs. Then the sparse high-resolution depth map is propagated to a dense depth map via a simple propagation module. Finally, a differentiable Gauss-Newton layer is proposed to optimize the depth map further. Experimental results on two challenging datasets verify the effectiveness and efficiency of the proposed method. 

%\TODO{change statement, dot direct copy}

%\TODO{future work, view intergation, view propagation, learned depth fusion}

\section*{Acknowledgements}
The work was supported by National Key R\&D Program of China (2018AAA0100704), NSFC \#61932020, and ShanghaiTech-Megavii Joint Lab.
We would like to thank Dongze Lian, Weixin Luo, Lei Jin and Shenhan Qian for their insightful comments during the preparation of the manuscript.
	
\newpage
\appendix
\section{Architecture}
As presented in the main paper, our Fast-MVSNet has three parts: sparse high-resolution depth map prediction, depth map propagation, and Gauss-Newton refinement. For the sparse high-resolution depth map prediction, our network is similar to MVSNet~\cite{yao2018mvsnet} except that we build a sparse cost volume in spatial domain and use fewer virtual depth planes (\eg, 96). Therefore, we can obtain a sparse high-resolution depth map at much lower cost. For the depth map propagation module, we use a 10-layer convolutional network to prediction the weights $W$. We show the details of this network in Table~\ref{tab:prop}. For the Gauss-Newton refinement, we use a similar network architecture as propagation module to extract deep feature representations of the input images $\{I_i\}_{i=0}^N$. In particular, Conv\_4 and Conv\_7 as in Table~\ref{tab:prop} are first interpolated to the same size and then are concatenated as the deep feature representation.

\begin{table}[h]
	\centering
	\footnotesize
	\resizebox{.48\textwidth}{!}{
	\begin{tabular}{c|l|c}
		\toprule
		Name          & Layer    & Output Size               \\
		\midrule
		Input           &       & H$\times$W$\times$3                  \\ 
		\midrule
		Conv\_0           & ConvBR,K=3x3,S=1,F=8       & H$\times$W$\times$8                  \\ 
		Conv\_1           & ConvBR,K=3x3,S=1,F=8                 & H$\times$W$\times$ 8                  \\ 
		Conv\_2           & ConvBR,K=5x5,S=2,F=16                & \sfrac{1}{2}H$\times$\sfrac{1}{2}W$\times$16         \\ 
		Conv\_3           & ConvBR,K=3x3,S=1,F=16                & \sfrac{1}{2}H$\times$\sfrac{1}{2}W$\times$16         \\ 
		Conv\_4           & ConvBR,K=3x3,S=1,F=16                & \sfrac{1}{2}H$\times$\sfrac{1}{2}W$\times$16         \\ 
		Conv\_5           & ConvBR,K=5x5,S=2,F=32                & \sfrac{1}{4}H$\times$\sfrac{1}{4}W$\times$32         \\ 
		Conv\_6           & ConvBR,K=3x3,S=1,F=32                & \sfrac{1}{4}H$\times$\sfrac{1}{4}W$\times$32         \\ 
		Conv\_7  & Conv,K=3x3,S=1,F=32                   & \sfrac{1}{4}H$\times$\sfrac{1}{4}W$\times$32      \\ 
		Conv\_8  & Conv,K=3x3,S=1,F=16                   & \sfrac{1}{4}H$\times$\sfrac{1}{4}W$\times$16      \\ 
		\midrule
		$W$  & Conv,K=3x3,S=1,F=$k^2$                  & \sfrac{1}{4}H$\times$\sfrac{1}{4}W$\times$ $k^2$      \\ 
		\bottomrule
	\end{tabular}
    }
	\centering
	\caption{Weights prediction network in the propagation module. We denote the 2D convolution as Conv and use BR to abbreviate the batch normalization and the Relu. K is the kernel size, S the kernel stride and F the output channel number. H, W denote image height and width, respectively.}
	\label{tab:prop}
\end{table}

\section{Depth maps fusion}
The fusion has three steps: photometric filtering, geometric consistency, and depth fusion. For photometric filtering, we first interpolate the predicted probability of the sparse high-resolution depth map to a high-resolution probability map and filter out points whose probability is below a threshold. The filtering threshold is set to 0.5. For geometric consistency, we compute the discrepancy of each depth map and filter out points whose discrepancy is larger than a threshold $\eta$. Specifically, a point $p$ in reference dpeth map $D$ is first projected to $p'$ in the neighboring depth map $\hat{D}$, then the discrepancy is defined as $f\cdot baseline \cdot \|\frac{1}{D(p)} -\frac{1}{\hat{D}(p')} \|$, where $f$ is the focal length of reference image and $baseline$ is the baseline of two images. The threshold $\eta$ is set to 0.12 pixels. For depth fusion, we require each point to be visible in $V = 3$ views and take the average value of all reprojected depths. 

\begin{figure*}[t!]
	\centering
	\footnotesize
	\includegraphics[width=1.0\textwidth]{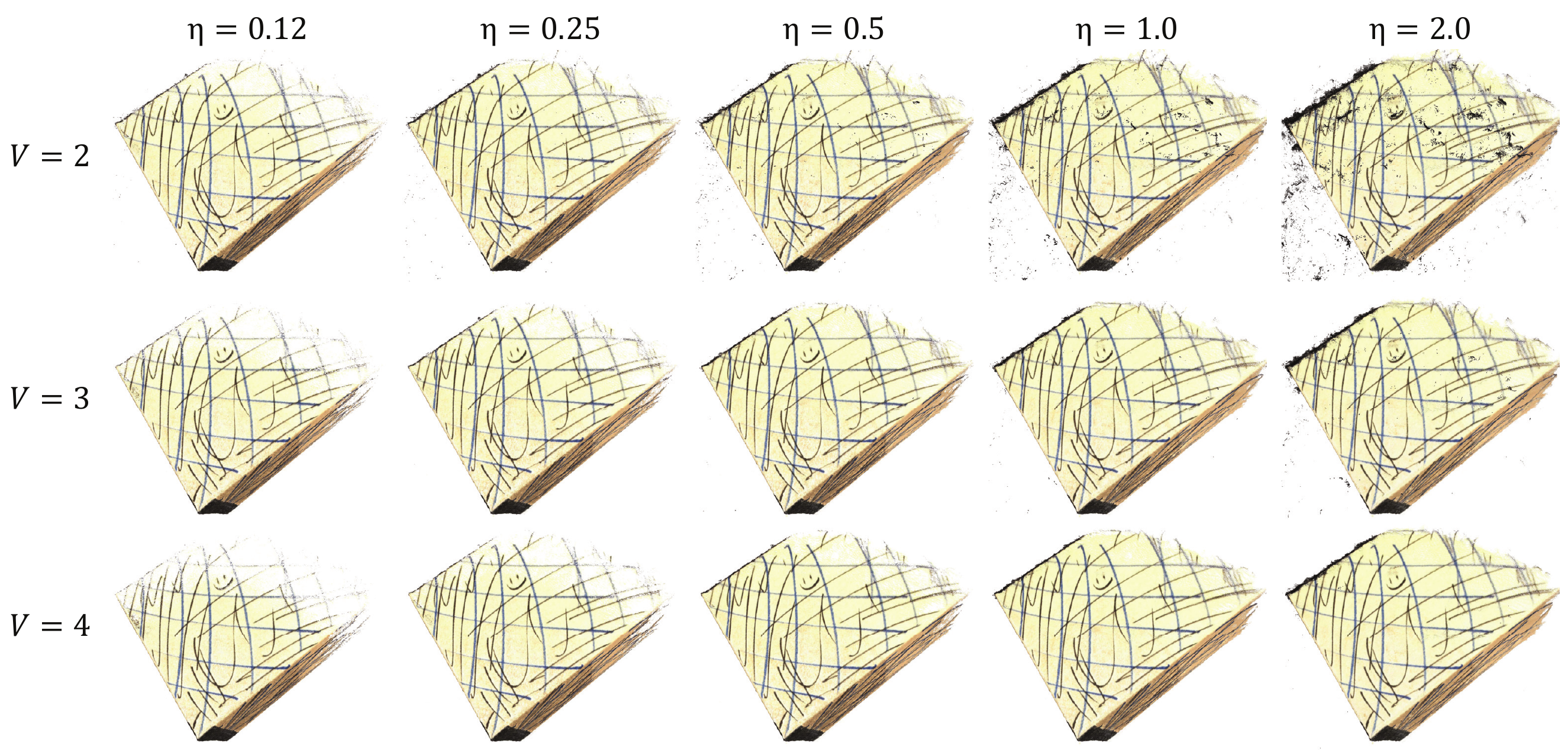}
	\caption{Reconstruction results of \textit{scan10} on the DTU dataset~\cite{aanaes2016large} with different fusion parameters. $\eta$ is the threshold of geometric consistency check. $V$ is the number of views that a point should be visible. As $\eta$ increases, the reconstruction becomes denser while has more noise. As $V$ increases, the reconstruction becomes cleaner while also becomes sparser. \label{fig:fusion}}
\end{figure*}

In the main paper, for a fair comparison, we use the same parameters for depth map fusion as that in Point-MVSNet~\cite{ChenPMVSNet2019ICCV}. However, we find that the fusion parameters $\eta$ and $V$ have a significant impact on reconstruction results. We show the quantitative comparison of reconstructions with different $\eta$ and $V$ in Table~\ref{tab:fuse}. The comparison of visualization results are shown in Figure~\ref{fig:fusion}. From the comparison results, we can see the trade off between \textit{Accuracy} and \textit{Completeness}. Increasing $\eta$, the reconstructed points gets less accurate but more complete. Increasing $V$, the reconstructions become more accurate while become incomplete. As the fusion has significant impact on the final reconstruction results, integrating a learnable fusion module~\cite{Donne_2019_CVPR} into the overall pipeline will be an interesting direction in future work.

\begin{table}[t]
	\centering
	\setlength{\belowcaptionskip}{-0.3cm}
	\footnotesize
	\begin{tabular}{cc|ccc}
		\toprule
		$\eta$ & $V$ & Acc. (mm)    & Comp. (mm) & Overall (mm) \\
		\midrule
		0.12 & 2  & 0.3969 & 0.3140 & \textbf{0.3555}   \\
		0.12 & 3  & 0.3360 & 0.4030 & 0.3695  \\
		0.12 & 4  & \textbf{0.3007} & 0.5212 & 0.4109  \\
		\midrule
		0.25 & 2  & 0.4663 & 0.2843 & 0.3753  \\
		0.25 & 3  & 0.3951 & 0.3341 & 0.3646  \\
		0.25 & 4  & 0.3542 & 0.3959 & 0.3750  \\
		\midrule
		0.5 & 2  & 0.5480 & \textbf{0.2773} & 0.4127  \\
		0.5 & 3  & 0.4614 & 0.3076 & 0.3845  \\
		0.5 & 4  & 0.4128 & 0.3447 & 0.3788  \\
		\midrule
		1.0 & 2  & 0.6655 & 0.2888 & 0.4772  \\
		1.0 & 3  & 0.5555 & 0.3091 & 0.4323  \\
		1.0 & 4  & 0.4923 & 0.3330 & 0.4126  \\
		\midrule
		2.0 & 2  & 0.8381 & 0.3187 & 0.5784  \\
		2.0 & 3  & 0.7002 & 0.3323 & 0.5163  \\
		2.0 & 4  & 0.6152 & 0.3500 & 0.4826  \\
		\bottomrule      
	\end{tabular}
	\caption{Quantitative results of reconstruction quality on the DTU evaluation dataset~\cite{aanaes2016large}. Increasing the geometric consistency threshold $\eta$, the reconstruted points become less accurate but also become more complete. Increasing the number of visible views $V$, the reconstruction becomes accurate while also becomes incomplete.}
	\label{tab:fuse}
\end{table}

\section{Gauss-Newton refinement with more iterations}
\begin{table}[t]
	\centering
	\footnotesize
	\begin{tabular}{c|ccc}
		\toprule
		\# iterations & Acc. (mm)    & Comp. (mm) & Overall (mm) \\
		\midrule
		 0  & 0.3679 & 0.4475 & 0.4077 \\
		 1  & \textbf{0.3360} & 0.4030 & 0.3695 \\
		 2  & 0.3391 & 0.3956 & 0.3673 \\
		 3  & 0.3420 & 0.3902 & 0.3662 \\
		 4  & 0.3435 & 0.3885 & 0.3660 \\
		 5  & 0.3443 & \textbf{0.3875} & \textbf{0.3659} \\
		\bottomrule      
	\end{tabular}
	\caption{Quantitative results of reconstruction quality on the DTU evaluation dataset~\cite{aanaes2016large} with different iteration number in Gauss-Newton refinement.}
	\label{tab:iter}
\end{table}

In this section, we conduct ablation study for Gauss-Newton refinement with more iterations. As shown in Table~\ref{tab:iter}, Gauss-Newton refinement can significantly improves the reconstruction quality. However, the performance improvements of applying Gauss-Newton refinement with more interations are marginal. Therefore, we only use one iteration in Gauss-Newton refinement.

\section{Reconstruction results}
\begin{figure*}[t!]
	\centering
	\footnotesize
	\includegraphics[width=0.87\textwidth]{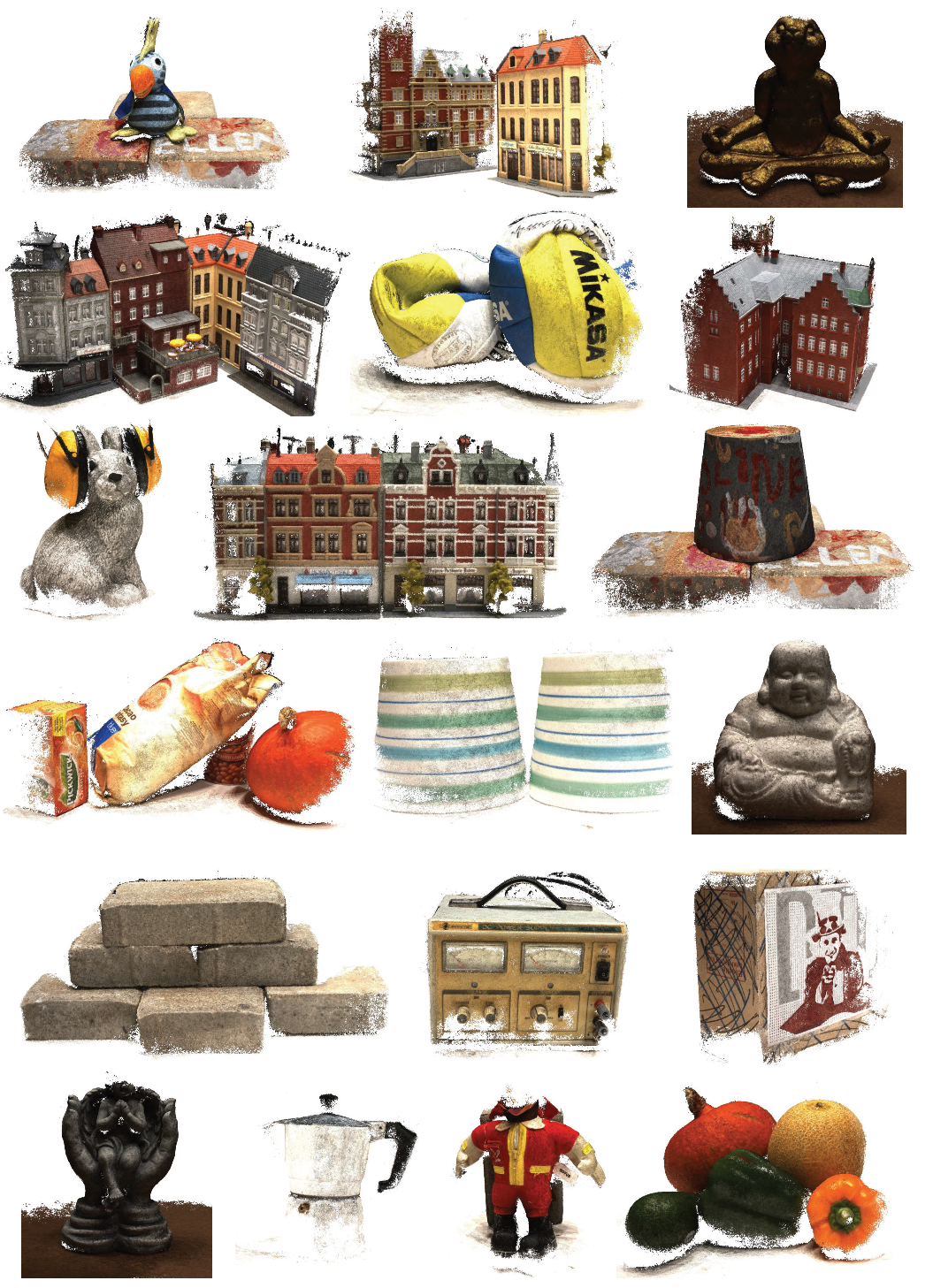}
	\caption{Reconstruction results on the DTU dataset~\cite{aanaes2016large}.\label{fig:all}}
\end{figure*}

We show more reconstruction results on the DTU dataset~\cite{aanaes2016large} in Figure~\ref{fig:all}. Our reconstruction is dense and accurate for all scenes.
{\small
	\bibliographystyle{ieee}
	\bibliography{library}
}	
\end{document}